\documentclass[journal]{IEEEtran}
\usepackage{cite}
\usepackage{times}
\usepackage{amsmath,graphicx}
\usepackage{epstopdf}
\usepackage{xcolor}
\usepackage{amssymb}
\usepackage{array}
\usepackage{booktabs}
\usepackage{multirow}
\usepackage{graphicx}
\usepackage{color}
\usepackage{float}
\usepackage{stfloats}
\usepackage[misc]{ifsym}
\usepackage{stmaryrd}
\usepackage{grffile}
\usepackage[caption=false,font=footnotesize]{subfig}
\usepackage{makecell}
\usepackage{float}
\usepackage{threeparttable}
\usepackage{enumitem}
\usepackage{setspace}
\usepackage[implicit=false]{hyperref}

\begin{document}

\title{Deformable 3D Convolution for Video Super-Resolution}

\author{Xinyi~Ying, Longguang~Wang, Yingqian~Wang, Weidong~Sheng, Wei An, and Yulan~Guo

\thanks{X.~Ying, L.~Wang, Y.~Wang, W.~Sheng, W.~An, and Y.~Guo are with the College of Electronic Science and Technology, National University of Defense Technology, P. R. China. Corresponding author: W.~Sheng. Emails: yingxinyi18@nudt.edu.cn, shengweidong1111@sohu.com, yulan.guo@nudt.edu.cn.
This work was partially supported by the National Natural Science Foundation of China (61605242, 61972435).}}

\markboth{Journal of \LaTeX\ Class Files,~Vol.~14, No.~8, August~2015}%
{Shell \MakeLowercase{\textit{et al.}}: Bare Demo of IEEEtran.cls for IEEE Journals}

\maketitle

\begin{abstract}
 The spatio-temporal information among video sequences is significant for video super-resolution (SR). However, the spatio-temporal information cannot be fully used by existing video SR methods since spatial feature extraction and temporal motion compensation are usually performed sequentially. In this paper, we propose a deformable 3D convolution network (D3Dnet) to incorporate spatio-temporal information from both spatial and temporal dimensions for video SR. Specifically, we introduce deformable 3D convolution (D3D) to integrate deformable convolution with 3D convolution, obtaining both superior spatio-temporal modeling capability and motion-aware modeling flexibility. Extensive experiments have demonstrated the effectiveness of D3D in exploiting spatio-temporal information. Comparative results show that our network achieves state-of-the-art SR performance. Code is available at: \url{https://github.com/XinyiYing/D3Dnet}.
\end{abstract}

\begin{IEEEkeywords}
Video super-resolution, deformable convolution.
\end{IEEEkeywords}

\section{Introduction}\label{sec1}

\IEEEPARstart{V}{ideo} super-resolution (SR) aims at recovering high-resolution (HR) images from low-resolution (LR) video sequences. This technique has been widely employed in many applications such as video surveillance \cite{surveillance1} and high-definition devices \cite{high1,high2}. Since video frames provide additional information in temporal dimension, it is important to fully use the spatio-temporal dependency to enhance the performance of video SR.

Current video SR methods commonly follow a three-step pipeline, which consists of feature extraction, motion compensation and reconstruction. Existing video SR methods generally focus on the motion compensation step and propose different approaches. Specifically, Liao \textit{et al.} \cite{draft} first achieved motion compensation using several optical flow algorithms to generate SR drafts and then ensembled these drafts via a CNN. Liu \textit{et al.} \cite{temporal_dynamic1,temporal_dynamic2} first performed rectified optical flow alignment and then fed these aligned LR frames to a temporal adaptive neural network to reconstruct an SR frame in an optimal temporal scale. Wang \textit{et al.} \cite{SOFVSR18,SOFVSR20} proposed an \textit{SOF-VSR} network to obtain temporally consistent details through HR optical flow estimation. Caballero \textit{et al.} \cite{VESPCN} proposed a spatial transformer network by employing spatio-temporal \textit{ESPCN} \cite{ESPCN} to recover an HR frame from compensated consecutive sequence in an end-to-end manner. Tao \textit{et al.} \cite{DRVSR} integrated a sub-pixel motion compensation (SPMC) layer into CNNs to achieve improved performance. All these methods perform motion compensation in two separate steps: motion estimation by optical flow approaches and frame alignment by warping, resulting in ambiguous and duplicate results \cite{Devon, TGA-VSR}.

 To achieve motion compensation in a unified step, Tian \textit{et al.} \cite{TDAN} proposed a temporally deformable alignment network (\textit{TDAN}) for video SR. Specifically, neighboring frames are first aligned to the reference frame by deformable convolution. Afterwards, these aligned frames are fed to CNNs to generate SR results. Wang \textit{et al.} \cite{EDVR} proposed an enhanced deformable video restoration network, namely \textit{EDVR}. The pyramid, cascading and deformable (PCD) alignment module of \textit{EDVR} can handle complicated and long-range motion and therefore improves the performance of video SR. Xiang \textit{et al.} \cite{zooming} proposed a deformable ConvLSTM method to exploit superior temporal information for video sequences with large motion. However, these aforementioned methods are two-stage methods. That is, feature extraction is first performed within the spatial domain and motion compensation is then performed within the temporal domain. Consequently, the spatio-temporal information within a video sequence cannot be jointly exploited and the coherence of the super-resolved video sequences is weakened.

Since 3D convolution (C3D) \cite{C3D} can model appearance and motion simultaneously, it is straightforward to apply C3D for video SR. Li \textit{et al.} \cite{FSTRN} proposed a one-stage approach (\textit{i.e.}, fast spatio-temporal residual network (\textit{FSTRN})) to perform feature extraction and motion compensation jointly. However, due to its fixed receptive field, C3D cannot model large motion effectively. To obtain both spatio-temporal modeling capability and motion-aware modeling flexibility, we integrate deformable convolution \cite{DCN1} with C3D  to achieve deformable 3D convolution (D3D). Different from the 3D deformable convolution in \cite{MRI} which was used for high-level classification task, our D3D is designed for low-level SR task and only perform kernel deformation in spatial dimension to incorporate the temporal prior (\textit{i.e.}, frames temporally closer to the reference frame are more important \cite{EDVR,TGA-VSR}) and reduce computational cost.

\begin{figure*}[t]
\centering
\vspace{-.15in}
\includegraphics[width=16.5cm]{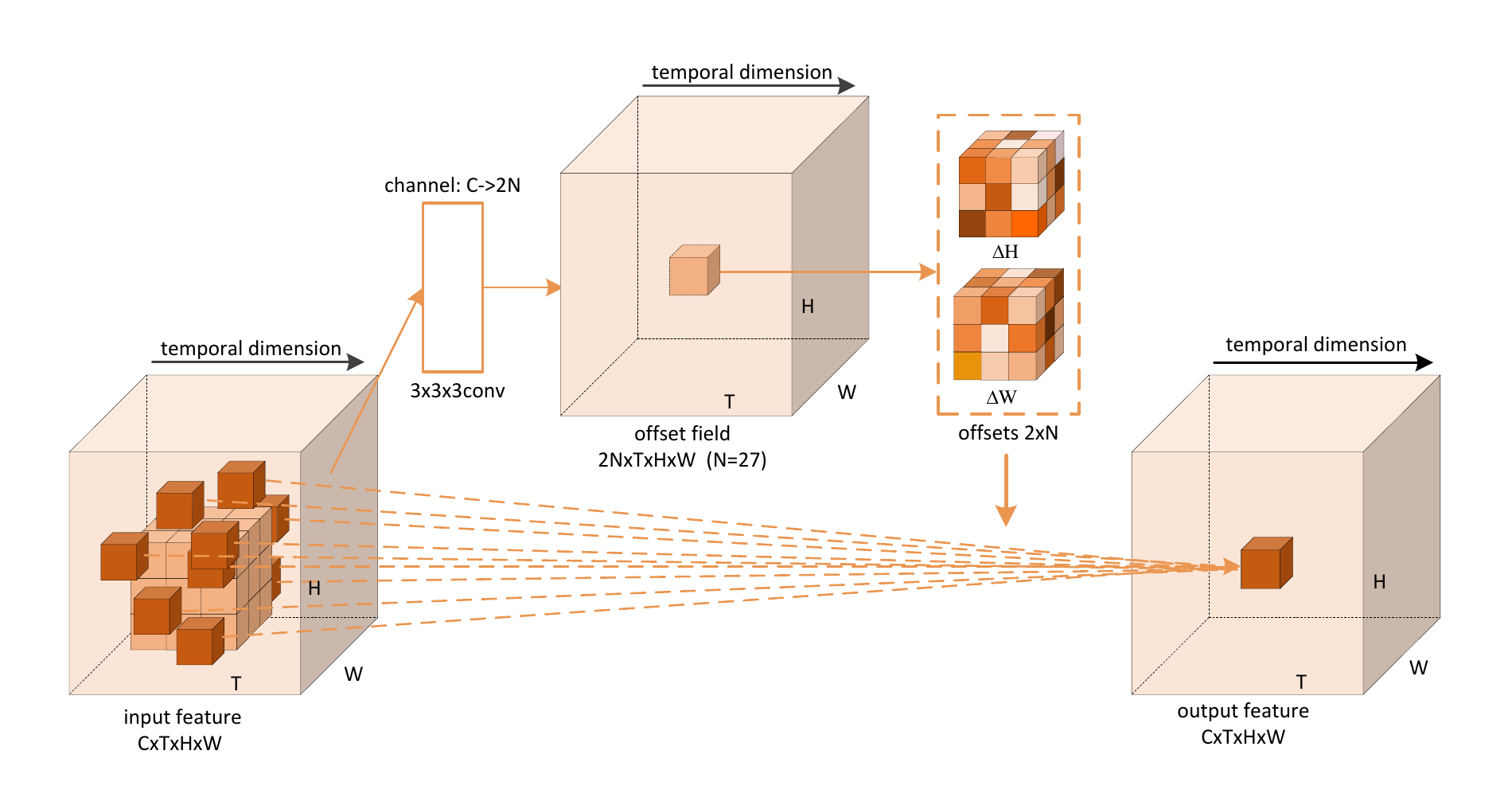}
\caption{A toy example of D3D. In the input feature of size $C\times T\times W\times H$, the light orange cubes represent the sampling grid of a plain 3$\times$3$\times$3 convolution, and the dark orange cubes represent the sampling grid of a deformable 3$\times$3$\times$3 convolution. The offsets are generated by an offset generator (3$\times$3$\times$3 convolution) and have $2N$ values along their  channel dimension, which represent the deformation values of D3D sampling grid ($N$ in height and $N$ in width). Here, $N=27$ is the size of the sampling grid.}\label{FigSD3D}
\end{figure*}

In summary, the contributions of this paper are as follows: 1) We propose a deformable 3D convolution network (D3Dnet) to fully exploit the spatio-temporal information for video SR. 2) We integrate deformable convolution \cite{DCN1} and 3D convolution \cite{C3D} to propose deformable 3D convolution (D3D), which can achieve efficient spatio-temporal information exploitation and adaptive motion compensation. 3) Extensive experiments have demonstrated that our D3Dnet can achieve state-of-the-art SR performance with high computational efficiency.

\begin{figure}[t]
\vspace{-.1in}
\centering\includegraphics[width=8.7cm]{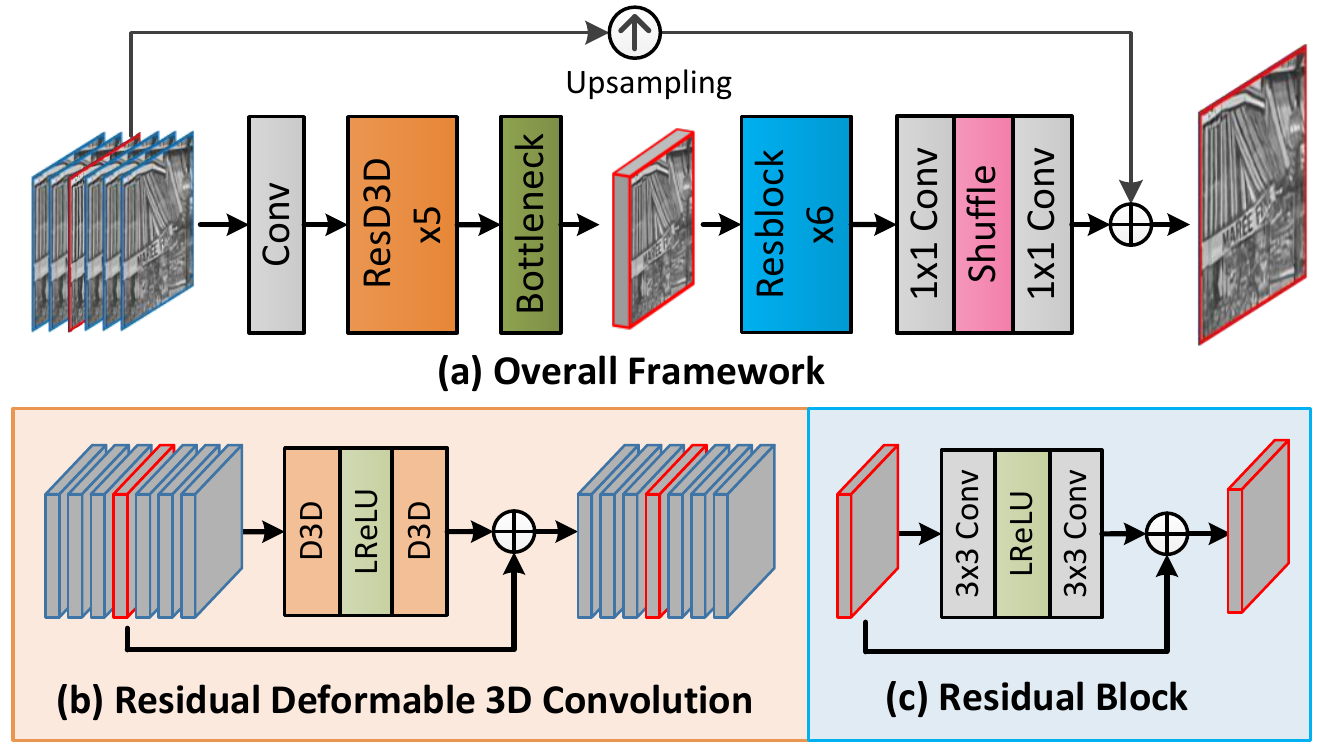}
\caption{An illustration of our D3Dnet. (a) The overall framework. (b) The residual deformable 3D
convolution (resD3D) block for simultaneous appearance and motion modeling. (c) The residual block for SR reconstruction.}\label{FigNetwork}
\end{figure}

\section{Methodology}\label{Methodology}

\subsection{Deformable 3D Convolution}\label{SecD3D}

 The plain C3D  \cite{C3D} is achieved in the following two steps: 1) 3D convolution kernel sampling on input features $x$, and 2) weighted summation of sampled values by function $w$. To be specific, the features passed through a regular 3$\times$3$\times$3 convolution kernel with a dilation of 1 can be formulated as:
\begin{align}\label{eq1}
y(\mathop p\nolimits_0 ) = \sum\limits_{n  = 1}^N {w(\mathop p\nolimits_n ) \cdot x(\mathop p\nolimits_0  + \mathop p\nolimits_n )},
\end{align}
where $\mathop p\nolimits_0$  represents a location in the output feature and $\mathop p\nolimits_n$ represents the $n_{th}$ value in 3$\times$3$\times$3 convolution sampling grid $\mathop G=\left\{ {\left( { - 1, - 1, - 1} \right),\left( { - 1, - 1,0} \right),...,\left( {1,1,0} \right),\left( {1,1,1} \right)} \right\}$. Here, $N=27$ is the size of the sampling grid. As shown in Fig.~\ref{FigSD3D}, the 3$\times$3$\times$3 light orange cubes in the input feature can be considered as the plain C3D sampling grid, which is used to generate the dark orange cube in the output feature.

Modified from C3D, D3D can enlarge the spatial receptive field with learnable offsets, which improves the appearance and motion modeling capability. As illustrated in Fig.~\ref{FigSD3D}, the input feature of size $C\times T\times W\times H$ is first fed to C3D to generate offset features of size $2N\times T\times W\times H$. Note that, the number of channels of these offset features is set to $2N$ for 2D spatial deformations (\textit{i.e.,} deformed along height and width dimensions). Then, the learned offsets are used to guide the deformation of the plain C3D sampling grid (\textit{i.e.}, the light orange cubes in the input feature) to generate a D3D sampling grid (\textit{i.e.}, the dark orange cubes in the input feature). Finally, the D3D sampling grid is employed to produce the output feature. In summary, D3D is formulated as:
\begin{align}\label{eq2}
y(\mathop p\nolimits_0) =  \sum\limits_{n  = 1}^{N} {w(\mathop p\nolimits_{n} ) \cdot x(\mathop p\nolimits_0  + \mathop p\nolimits_{n} + \mathop {\Delta p}\nolimits_{n})},
\end{align}
where $\mathop {\Delta p}\nolimits_{n}$ represents the offset corresponding to the $n_{th}$ value in  3$\times$3$\times$3 convolution sampling grid.
Since the offsets are generally fractional, we followed \cite{DCN1, DCN2} to use bilinear interpolation to generate exact values.


\begin{figure}[t]
\vspace{-.1in}
\centering\includegraphics[width=8.7cm]{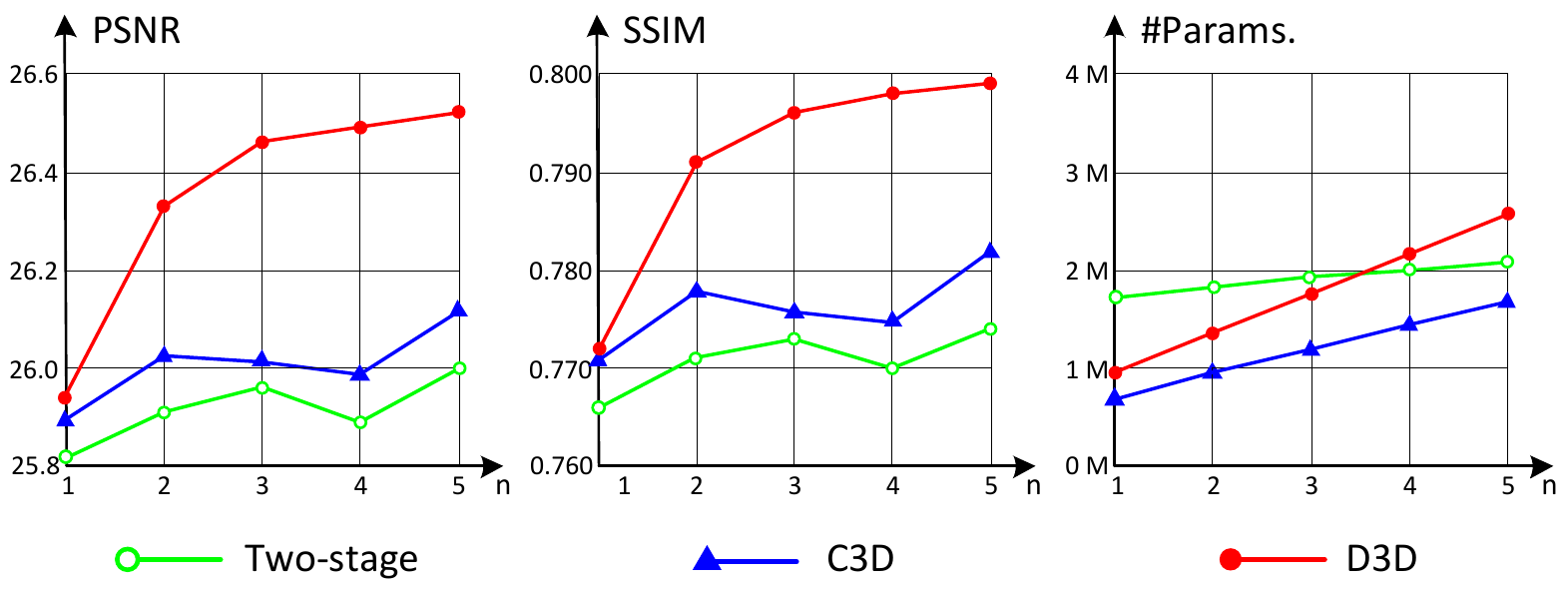}
\caption{Comparative results achieved on Vid4 \cite{VESPCN} dataset of the two-stage method and one-stage methods (C3D and D3D) with respect to the number of residual blocks. ``\#Params." represents the number of parameters.}\label{FigAblation}
\end{figure}

\subsection{Overall Framework}\label{SecD3D}

The overall framework is shown in Fig.~\ref{FigNetwork}(a). Specifically, a video sequence with 7 frames is first fed to a C3D layer to generate features, which are then fed to 5 residual D3D (resD3D) blocks (shown in Fig.~\ref{FigNetwork}(b)) to achieve motion-aware deep spatio-temporal feature extraction. Then, a bottleneck layer is employed to fuse these extracted features. Finally, the fused feature is processed by 6 cascaded residual blocks (shown in Fig. 2(c)) and a sub-pixel layer for SR reconstruction. We use the mean square error (MSE) between the super-resolved and the groundtruth reference frame as the training loss of our network.

\section{Experiments}\label{sec4}

\subsection{Implementation Details}

For training, we employed the Vimeo-90k dataset \cite{Vimeo} as the training set with a fixed resolution of 448$\times$256. To generate training data, all video sequences were bicubically downsampled by 4 times to produce their LR counterparts. Then, we randomly cropped these LR images into patches of size $32\times32$ as input. Their corresponding HR images were cropped into patches accordingly. We followed \cite{SOFVSR20,SOFVSR18} to augment the training data by random flipping and rotation.

For test, we employed the Vid4 \cite{VESPCN}, Vimeo-90k \cite{Vimeo} and SPMC \cite{DRVSR} datasets for performance evaluation. Following \cite{TDAN,PASSRnet,LF-InterNet}, we used peak signal-to-noise ratio (PSNR) and structural similarity index (SSIM) as quantitative metrics to evaluate SR performance. In addition, we used motion-based video integrity evaluation index (MOVIE) and temporal MOVIE (T-MOVIE) \cite{MOVIE} to evaluate temporal consistency. All metrics were computed in the luminance channel.

 All experiments were implemented in Pytorch with an Nvidia RTX 2080Ti GPU. The networks were optimized using the Adam method \cite{Adam}. The learning rate was initially set to $4\times10^{-4}$ and halved for every 6 epochs. We stopped the training after 35 epochs.

\subsection{Ablation Study}\label{ablation}

\subsubsection{One-stage vs. Two-stage}\label{fine-tune}
We designed two variants to test the performance improvement introduced by integrating feature extraction and motion compensation. For the two-stage variant, we replaced the resD3D blocks with $n$ residual blocks and deformable alignment module \cite{TDAN} to sequentially perform spatial feature extraction and temporal motion compensation. For the one-stage variant, we replaced resD3D blocks with $n$ residual C3D blocks to integrate the two steps without spatial deformation. It can be observed in Fig.~\ref{FigAblation} that the PSNR and SSIM values of the two-stage variant are lower than the one-stage variant (\textit{i.e.,} C3D) by 0.10 and 0.006 in average. Moreover, the two-stage variant has more parameters than the one-stage variant. It demonstrates that the one-stage method can fully exploit the spatio-temporal information for video SR with fewer parameters.

\begin{figure*}[t]
\centering
\vspace{-.1in}
\includegraphics[width=17.9cm]{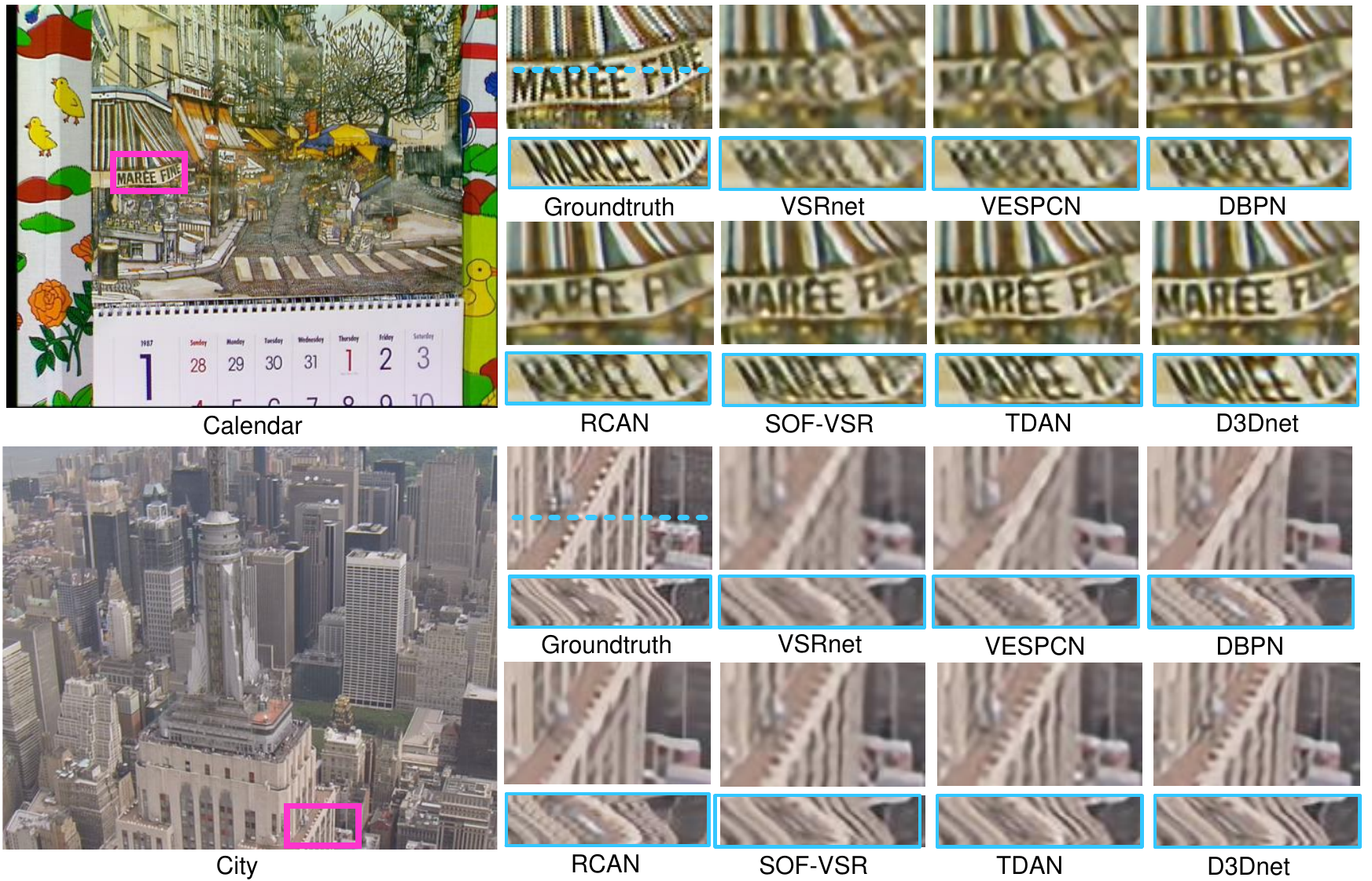}
\caption{Qualitative results achieved by different methods. Blue boxes represent the temporal profiles among different frames.}\label{visual}
\vspace{0.2cm}
\end{figure*}

\begin{table}
\footnotesize
\centering
\renewcommand\arraystretch{1.5}
\caption{Results achieved on Vid4 \cite{VESPCN} dataset by D3Dnet trained with different number of input frames (\textit{i.e.,} ``Fra.'')}\label{Frames}
\scriptsize
\begin{tabular}{c|cccc|c}
\hline
Fra. & City&Walk&Calendar &Foliage  &Average\\
\hline
$3$&27.00/0.765&29.31/0.889&22.98/0.762& 25.61/0.729 &26.22/0.786\\
$5$&27.16/0.776&29.63/0.895&23.19/0.773& 25.79/0.740 &26.44/0.796\\
$7$&27.23/0.780&29.72/0.896&23.26/0.775& 25.88/0.745 &26.52/0.799\\
\hline
\end{tabular}\label{InputFrame}
\end{table}

\subsubsection{C3D vs. D3D}
To test the performance improvement introduced by the deformation operation, we compare the results of our network with different number of residual C3D (resC3D) blocks and resD3D blocks. It can be observed in Fig.~\ref{FigAblation} that D3D achieves significant performance improvement over C3D. Specifically, our network with 5 resD3D blocks achieves an improvement of 0.40 dB in PSNR and 0.017 in SSIM as compared to the network with 5 resC3D blocks. Note that, due to the offset generation branch, each resD3D block introduces 0.19M additional parameters.

\begin{table*}
\footnotesize
\centering
\renewcommand\arraystretch{1.4}
\caption{Quantitative results (PSNR/SSIM) of different methods achieved on Vid4 \cite{VESPCN}, Vimeo-90k \cite{Vimeo} and SPMC-11 \cite{DRVSR} datasets. Best results are shown in boldface.}\label{quantitative1}
\scriptsize
\begin{tabular}{|l|c|c|c|c|c|c|c|c|c|c|}
\hline
Methods&Bicubic & \textit{VSRnet} \cite{VSRnet}&\textit{VESPCN} \cite{VESPCN}& \textit{DBPN} \cite{DBPN}&\textit{RCAN} \cite{RCAN}&\textit{SOF-VSR} \cite{SOFVSR20}&\textit{TDAN} \cite{TDAN}&\textit{D3Dnet}\\
\hline
Vid4 \cite{VESPCN}& 23.76 / 0.631	&24.37 / 0.679 &24.95 / 0.714&	25.32 / 0.736&	25.46 / 0.740&	26.02 / 0.771&	26.16 / 0.782&	\textbf{26.52 / 0.799}\\

Vimeo-90k \cite{Vimeo}&31.31 / 0.865& 32.43 / 0.889	&	33.55 / 0.907	&35.17 / 0.925	& 35.35 / 0.925	&34.89 / 0.923	&35.34 / 0.930&	\textbf{35.65 / 0.933}\\

SPMC-11 \cite{DRVSR}&25.67 / 0.726& 26.41 / 0.766&	27.09 / 0.794&	27.92 / 0.822	&28.36 / 0.828	&28.21 / 0.832	&28.51 / 0.841	& \textbf{28.78 / 0.851}\\
\hline
Average &26.91 / 0.741&27.74 / 0.778&	28.53 / 0.805	&29.47 / 0.828	&29.72 / 0.831	&29.71 / 0.842	&30.00 / 0.851	&\textbf{30.32 / 0.861}\\
\hline
\end{tabular}
\end{table*}

\subsubsection{Context Length}\label{ablationNums}
 The results of our D3Dnet with different number (\textit{i.e.}, 3, 5 and 7) of input frames are shown in Table~\ref{InputFrame}. It can be observed that the performance improves as the number of input frames increases. Specifically, the PSNR/SSIM improves from 26.22/0.786 to 26.52/0.799 when the number of input frames increases from 3 to 7. That is because, more input frames introduce additional temporal information, which is beneficial for video SR.

\subsection{Comparison to the State-of-the-arts}

In this section, we compare our D3Dnet with 2 single image SR methods (\textit{i.e.}, \textit{DBPN} \cite{DBPN} and \textit{RCAN} \cite{RCAN}) and 4 video SR methods (\textit{i.e.}, \textit{VSRnet} \cite{VSRnet}, \textit{VESPCN} \cite{VESPCN},  \textit{SOF-VSR} \cite{SOFVSR20, SOFVSR18}, and \textit{TDAN} \cite{TDAN}). We also present the results of bicubic interpolation as the baseline results. Note that, \textit{EDVR} \cite{EDVR} and \textit{DUF-VSR} \cite{DUF-VSR} are  not included in our comparison due to the large gap in computational cost. Specifically, the \#Params$/$FLOPs of our D3Dnet are 12.5\%$/$17.0\% of those of \textit{EDVR} \cite{EDVR} and 44.5\%$/$25.4\% of those of \textit{DUF-VSR}  \cite{DUF-VSR}. For fair comparison, the first and the last 2 frames of the video sequences were excluded for performance evaluation.

Quantitative results are listed in Tables~\ref{quantitative1} and \ref{quantitative2}. D3Dnet achieves the highest PSNR and SSIM scores among all the compared methods. That is because, D3D improves the spatial information exploitation capability and perform motion compensation effectively. In addition, D3Dnet outperforms existing methods in terms of MOVIE and T-MOVIE by a notable margin, which means that the results generated by D3Dnet are temporally more consistent.

\begin{table}
\footnotesize
\centering
\renewcommand\arraystretch{1.2}
\caption{Temporal consistency and computational efficiency achieved on Vid4 \cite{VESPCN} dataset. ``Fra." represents the number of input frames. ``\#Params." represents the number of parameters. FLOPs is computed based on HR frames with a resolution of 1280$\times$720.
Best results are shown in boldface.}\label{quantitative2}
\scriptsize
\begin{tabular}{|p{1.3cm}|c|c|c|c|c|c|}
\hline
Methods&Fra.&T-MOVIE&MOVIE&\#Params.&FLOPs&Time\\
\hline
\textit{DBPN} \cite{DBPN}&1&21.43&5.50&10.43M&5213.0G&34.9s\\
\textit{RCAN} \cite{RCAN}&1&23.49&5.98&15.59M&919.20G&134.5s\\
\textit{VSRnet} \cite{VSRnet}&5&26.05&6.01&\textbf{0.27M}&260.88G&46.2s\\
\textit{VESPCN} \cite{VESPCN}&3&25.41&6.02&0.88M&\textbf{49.83G}&23.6s\\
\textit{SOF-VSR} \cite{SOFVSR20}&3&19.35&4.25&1.64M&108.90G&\textbf{12.6s}\\
\textit{TDAN} \cite{TDAN}&5&18.87&4.11&1.97M&288.02G&30.6s\\
\textit{D3Dnet}&7&\textbf{15.45}&\textbf{3.38}&2.58M&408.82G&45.2s\\
\hline
\end{tabular}
\end{table}

Qualitative results are shown in Fig.~\ref{visual}. It can be observed from the zoom-in regions that D3Dnet can recover finer details (\textit{e.g.}, the sharp edge of the word `MAREE' and the clear and smooth roof pattern). In addition, the temporal profiles of D3Dnet are clearer and smoother than other methods. That is, our network can produce visual pleasing results with higher temporal consistency. A demo video is available online at \url{https://wyqdatabase.s3-us-west-1.amazonaws.com/D3Dnet.mp4}.

\subsection{Computational Efficiency}

The computational efficiency (the number of parameters, FLOPs, and running time) are evaluated in Table~\ref{quantitative2}. Note that, running time is the total time tested on the Vid4  \cite{VESPCN} dataset and is averaged over 20 runs. As compared with two single image SR methods \cite{DBPN,RCAN}, our D3Dnet achieves improvements in both SR performance and computational efficiency. Specifically, the number of parameters and FLOPs of D3Dnet are 16.5\% and 44.5\% of those of RCAN \cite{RCAN}. As compared with video SR methods \cite{VSRnet,VESPCN,SOFVSR20,TDAN}, our D3Dnet achieves better SR performance with a reasonable increase in computational cost.

\section{Conclusion}\label{sec5}
In this paper, we have proposed a deformable 3D convolution network (D3Dnet) to exploit spatio-temporal information for video SR. Our network introduces deformable 3D convolutions (D3D) to model appearance and motion simultaneously. Experimental results have demonstrated that our D3Dnet can effectively use the additional temporal information for video SR and achieve the state-of-the-art SR performance.

\bibliographystyle{IEEEtran}
\bibliography{D3Dnet}

\end{document}